\documentclass{article}

\usepackage[final]{neurips_2021}
\usepackage[pdftex]{graphicx}

\usepackage[utf8]{inputenc} % allow utf-8 input
\usepackage[T1]{fontenc}    % use 8-bit T1 fonts
\usepackage{hyperref}       % hyperlinks
\usepackage{url}            % simple URL typesetting
\usepackage{booktabs}       % professional-quality tables
\usepackage{amsfonts}       % blackboard math symbols
\usepackage{nicefrac}       % compact symbols for 1/2, etc.
\usepackage{microtype}      
\usepackage{xcolor}         % colors
\usepackage{textcomp}
\usepackage{amsfonts}
\usepackage[mathscr]{eucal}
\usepackage{graphicx}
\usepackage{mathtools}
 
\usepackage{booktabs}
\usepackage{amssymb}
\usepackage{diagbox}
\usepackage{subcaption}
\usepackage{amssymb}% http://ctan.org/pkg/amssymb
\usepackage{pifont}% http://ctan.org/pkg/pifont
\newcommand{\cmark}{\ding{51}}%
\newcommand{\xmark}{\ding{55}}%

\usepackage{multirow}
\usepackage{pifont}
\usepackage{multicol}

\title{What is Lost in Knowledge Distillation?}

\author{Manas Mohanty\hspace{10mm}Tanya Roosta \hspace{10mm}Peyman Passban\\
\texttt{mhmanas@amazon.com}\\
}
\begin{document}
\maketitle
\begin{abstract}
Deep neural networks (DNNs) have improved NLP tasks significantly, but training and maintaining such networks could be costly. Model compression techniques, such as, knowledge distillation (KD), have been proposed to address the issue; however, the compression process could be lossy. Motivated by this, our work investigates how a distilled student model differs from its teacher, if the distillation process causes any information losses, and if the loss follows a specific pattern. Our experiments aim to shed light on the type of tasks might be less or more sensitive to KD by reporting data points on the contribution of different factors, such as the number of layers or attention heads. Results such as ours could be utilized when determining effective and efficient configurations to achieve optimal information transfers between larger (teacher) and smaller (student) models.
\end{abstract}

\section{Introduction}\label{intro}
DNNs have contributed significantly to enabling advances in many tasks, including NLP. In this context, \textit{deep} refers to a large number of neural parameters. With the advancements in the computing hardware, training of DNNs with billions of parameters is now feasible. However, it does not necessarily mean that utilizing them for different setups has become easier. The computational complexity, storage requirements, and added latency still make their deployment in real-world difficult.  This issue is more pronounced on the ever-smaller devices, with limited processing and storage capacity. 

To make DNNs useful for low-budget settings, there are a number of approaches have been developed for model compression \cite{DBLP:journals/corr/abs-1710-09282}, which fall under one of the following categories of \textit{Pruning} \citep{MLSYS2020_d2ddea18}, \textit{Quantization} \cite{DBLP:journals/corr/CourbariauxBD15,sindhwani2015structured,DBLP:journals/corr/WuLWHC15, DBLP:journals/corr/ZhaiCLZ16a}, or knowledge distillation (KD) \citep{DBLP:journals/corr/abs-1910-01108,DBLP:journals/corr/abs-2006-05525,DBLP:journals/corr/abs-2012-14022}. Our work focuses on KD in the hope of understanding what part of knowledge is transferred during distillation. Specifically, the research question we aim to answer is \textit{what type of information is lost during the distillation process?} It might not be possible to fully dissect this concept but it is defiantly worthwhile to understand what classes, sets of tasks, and parameters are impacted more than others during distillation, so that the right mitigation can be put in place to account for the loss.

\section{Background}\label{background}
KD refers to transferring knowledge from a bigger or deeper model, called teacher ($\mathcal{T}$), to a smaller or shallower model, called student ($\mathcal{S}$). The term was first coined in \citet{Bucila2006ModelC}, and then elaborated by \citet{hinton2015distilling}, where KD was presented as a compression framework. Compression is achieved by a student model imitating its teacher's output. The output of a DNN is usually a class probability: $$p(z_i,T)=\frac{\exp(z_i/T)}{\sum_j{\exp(z_j/T)}}$$
where $z_i$ is the \textit{i}-th class's logit and $T$ is called the temperature. The loss function for distillation is usually formulated as:$$
\mathcal{L}_{KD}=Loss(p(z_t,\mathcal{T}),p(z_s,\mathcal{S}))$$
where the \textit{Loss} function is often a KL divergence \cite{DBLP:journals/corr/abs-2006-05525}, and $z_s$ and $z_t$ are the logits for the student and teacher models, respectively. The total loss is then an interpolation of the task loss ($\mathcal{L}_{task}$), and the KD loss:$$ L = \alpha \times \mathcal{L}_{task} + \eta \times \mathcal{L}_{KD}
$$
with $\alpha$ and $\eta$ (usually $\eta = 1 - \alpha$) as the weight values to indicate the contribution of each term to the learning process. 

\section{Related Work}
According to \citet{DBLP:journals/corr/abs-2006-05525}, KD approaches can be categorized into the following types of: \textit{i}) Response-based KD, where the logits of the output layers are matched between the teacher and the student models \citep{DBLP:journals/corr/abs-1910-01108}; \textit{ii}) Feature-based KD, where the outputs of specific intermediate hidden layers as well as those of the final layers are used to train the student model \citep{sun2019patient,wu-etal-2020-skip}; and \textit{iii}) Relation-based KD which is similar to the feature-based approach as both the intermediate and final output layers are utilized in training, but instead of using specific layers, a relationship between different layers is explored \citep{DBLP:journals/corr/abs-2012-14022}. In this paper, we use the response-based and feature-based KD approaches in our experiments, which enable us to keep the setup more manageable to investigate the problem. Relation-based KD introduces multiple moving/tunable factors which makes it enigmatic to justify and understand if the losses captured in the experiments are rooted in KD or in layer-to-layer communications. 

The research on the topic of KD is of great interest to many, and there are myriad of prior work on various aspects of it. However, to the best of our knowledge, the following papers are the closest to our research point of view. \citet{DBLP:journals/corr/abs-1911-05248} looked at the question of what compressed neural networks forget. They ask the question of whether the test accuracy is the best measure to determine whether the compressed model can generalize. They looked at the cases where the top level performance metrics are similar between the base and compressed model; however, some of the classes are dis-proportionally impacted by the compression. They tried to understand what makes some of the classes more sensitive to performance degradation. Their work specifically focused on image recognition and is on the pruning and quantization sides, not KD, but the way they probed DNNs is conceptually similar to our investigating. 

\citet{DBLP:journals/corr/abs-1905-10650} looked at the attention-based models in NLP and tried to determine if all attention heads are necessary when making predictions. They made an observation that for a set of tasks, during inference time, having sixteen heads for attention is not necessarily better than having only one attention head. This form of investigation is also similar to what we carried out in our research. 

In \citet{minilm}, the authors proposed a KD method for Transformers which is task-agnostic. Their method does not require a certain architecture guarantee for the student model and showed improvements on various GLUE tasks. The task-independent nature of this work makes it suitable to investigate the behaviour of KD methods.  

To the best of our knowledge, there has been no work performed on determining how KD affects different inference tasks, and how it affects classes with different labels. In this paper, we focus on these research questions. We believe, this could be of great use since existing NLP models are typically large, but the real-world use cases dictate limited resources. Understanding what information is lost (or retained) for a given configuration in a given task, can help guide system designers optimize their setup.

\section{Models and Experimental Hypotheses}\label{setup}
For our experiments, we use a 12-layer RoBERTa base model \citep{DBLP:journals/corr/abs-1907-11692} as the teacher. It is a well-studied and widely used model in the NLP community, which makes the reproduction of our work easier for other researchers. In each distillation setup, we keep all the configuration parameters untouched apart from the one that is of interest, e.g. we fix all the configuration parameters except the number of hidden layers. Then, we distill the base model into different student models each with a different number of hidden layers. The same approach is used for all the other configuration parameters. This leads to training of 3L, 6L, 9L, 4AH, 8AH, 384D, 516D, and 6L\_384D students, where L, AH, and D stand for Layer, Attention Head, and Dimension, respectively, i.e. 9L refers to a model that is identical to its teacher but it only has 9 layers. 

When reducing the size of the teacher model (from 12 to fewer layers), we had to come up with a mapping strategy to carry out the layer-to-layer distillation. After investigating a comprehensive set of mappings, the solution we arrived at was to always connect the first and last layers of the student and teacher models in all configurations. For internal layers, in 6L, with the exception of the third layer we skipped every other teacher layer and connected the rest uniformly. For 3L, we connected the fifth layer of the teacher to the middle layer of the student. In 9L, we connected the first 2 and the last 5 layers of the student to the first 2 and last 5 layers of the teacher, and aligned the third and fourth student layers to the fourth and sixth teacher layers, respectively. Similar to the number of layers, we also investigated the impact of tighter layers by shrinking the widths of the internal units to $516$ and $384$. We evaluated the distilled models over 8 tasks from GLUE \citep{DBLP:journals/corr/abs-1804-07461}, using the standard training and development sets from the various GLUE tasks. 

In regard to hyper-parameter, for distilling \textit{all configurations}, the hard, intermediate, and KD loss weights are each equally set to $0.33$. The temperature is set to $2$ and each batch processes $12$ instances. For fine-tuning, the learning rate of $5e-5$ worked best. Adam \citep{kingma2014adam} with $\beta_1 = 0.9$ and $\beta_2 = 0.999$ was used as the optimizer, and the number of epoch for fine-tuning was $10$. 

\section{Experimental Results}\label{results}
Table \ref{tab:scores} summarizes all our observations. These numbers can be interpreted from different perspectives and can vary in different settings. However, our findings show that: SST2 and QQP are the most resistant tasks (datasets) in the presence of KD whereas RTE is the most sensitive one. CoLA is also impacted severely since it loses most of the information through the KD process, across all configurations. QNLI, MRPC, and STSB are only impacted slightly and behave similarly. MNLI is in the middle of these two extremes; KD impacts it to some extent but not as much as the sensitive group. Apart from MNLI, there is a direct relation between the size of the training set and the KD loss, i.e. the smaller the set, the higher the loss. However, MNLI breaks this pattern as it has the largest dataset among all but still shows relatively high losses. This could be due to the complex nature of the inference task. It is the only dataset with $3$ classes, compared to others with $\leq 2$ classes.
\begin{table*}[th]
\begin{center}
\begin{tabular}{@{}ccccccccc@{}}
 \toprule
    & CoLA & MNLI & MRPC & QNLI & QQP  & RTE  & SST2 & STSB \\ \midrule
\textit{baseline} & 0.62 & 0.88 & 0.92 & 0.93 & 0.9  & 0.8  & 0.95 & 0.91 \\
\textit{9L} & 0.59 & 0.83 & 0.89 & 0.9  & 0.89 & 0.68 & 0.93 & 0.87 \\
\textit{6L} & 0.6  & 0.83 & 0.87 & 0.9  & 0.88 & 0.69 & 0.91 & 0.88 \\
\textit{3L} & 0.4  & 0.77 & 0.81 & 0.85 & 0.83 & 0.58 & 0.91 & 0.81 \\
\textit{516D} & 0.55 & 0.81 & 0.87 & 0.88 & 0.89 & 0.62 & 0.9  & 0.87 \\
\textit{384D} & 0.53 & 0.8  & 0.88 & 0.87 & 0.87 & 0.65 & 0.9  & 0.86 \\
\textit{8AH} & 0.55 & 0.81 & 0.87 & 0.88 & 0.88 & 0.61 & 0.92 & 0.85 \\
\textit{4AH} & 0.55 & 0.81 & 0.88 & 0.88 & 0.89 & 0.61 & 0.92 & 0.85 \\
\textit{6L\_384D} & 0.47 & 0.78 & 0.85 & 0.84 & 0.87 & 0.62 & 0.91 & 0.84 \\ \bottomrule
\end{tabular}
\caption{\label{tab:scores} The performance scores of KD models. STSB and CoLA use Pearson-Spearman and Matthews correlation coefficients. MRPC and QQP use the average of Accuracy and F1 scores and others rely on accuracy as their evaluation metric.}
\end{center}
\end{table*}

Reducing the number of layers up to a certain threshold is the best form of KD with only marginal losses. Both 9L and 6L show acceptable losses while reducing the number of layers to 3 was a drastic change and impacted the model significantly. The loss introduced by modifying the width of the hidden layers depends on the nature of the task and the dataset, but our results indicate that it can still be considered as a reasonable KD alternative. The loss is defiantly larger than that of the layer reduction but still is in an acceptable range. Combining both ideas, namely reducing the number and width of layers together is harmful and causes serious deterioration. 

Different researchers \citep{Michel2019AreSH,DBLP:journals/corr/abs-2110-15225} have previously reported that the number of attention heads can be reduced with almost \textit{no significant impact} on final results. With this assumption, we were expecting marginal losses after changing the number of heads but it did not happen in our case. We noticed that the number of heads plays a critical role in KD.  

\begin{figure}[htp]
    \centering
    \includegraphics[width=0.5\textwidth]{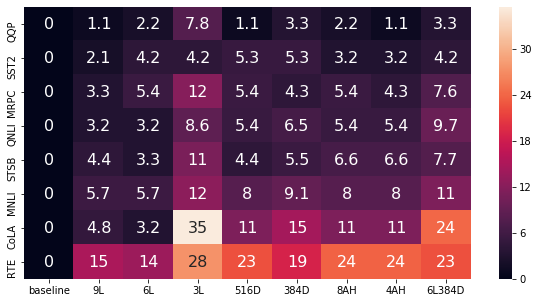}
    \caption{How much distillation degrades the student model. The intensity of the color of the cells (and numbers inside) show the percentage of disagreement between the student and teacher models, e.g. [CoLA][3L] = 35 indicates that 3L is 35$\%$ weaker than its teacher on the CoLA dataset.}
    \label{fig:heatmap}
\end{figure}

By investigating more data points and different aspects of the results, we hope to find a pattern for losses introduced by KD. We thus produced a heatmap of student-teacher disagreements in Figure \ref{fig:heatmap} based on the results from Table \ref{tab:scores}. As the figure shows, the highest losses belong to \textit{3L} models and more challenging datasets are \textit{RTE} and \textit{CoLA} on average. The heatmap simply visualized which tasks or models behave similarly.   
\subsection{Response Time}
KD is a compression strategy so we expect faster engines as the result of this process. To study how KD impacts models' response time we measured the number of samples per second each model can process under a common setting. Results of this experiment are reported in Table \ref{tab:speed}. We used a p3.8x environment with a set of four V100 GPUs (from AWS). Across all configurations and models we see a faster response time. However, students with fewer attention heads break this pattern. In theory, reducing the number of heads should also help with the speed increase whereas this is not observed in our experiments. Multiple factors, such as implementation techniques, hardware, non-optimal matrix multiplication etc can lead to this result, for example, in Transformers, parallelism mainly comes from the multi-head attention mechanism, where different heads can attend to different parts of the input sequence simultaneously. Each attention head processes a different subspace of the input, which allows for parallelization. Therefore, reducing the number of attention heads can reduce the scope of parallelism, which might lead to slow down in inference.      

\begin{table*}[th]
\begin{center}
\begin{tabular}{@{}lllllllll@{}}
\toprule
    & CoLA & MNLI  & MRPC  & QNLI  & QQP   & RTE   & SST2  & STSB  \\ \midrule
\textit{baseline} & 49.1 & 185.8 & 172   & 184.5 & 186.7 & 165.6 & 180.6 & 181.3 \\
\textit{9L} & 50.4 & 184.2 & 175.8 & 183.7 & 185.3 & 170.6 & 182.5 & 183.8 \\
\textit{6L} & 56   & 278.1 & 262.3 & 277.5 & 280.4 & 260.5 & 260.1 & 275.5 \\
\textit{3L} & 57.3 & 324.2 & 305   & 321.2 & 324.5 & 301.9 & 318.4 & 316.4 \\
\textit{516D} & 51.9 & 210.4 & 195.6 & 210.1 & 211.1 & 191   & 205.6 & 208.6 \\
\textit{384D} & 54   & 262   & 240.8 & 260.3 & 265.2 & 231.5 & 253.5 & 256.9 \\
\textit{8AH} & 46.9 & 152.7 & 139.3 & 153.6 & 154.2 & 140.5 & 150.8 & 153   \\
\textit{4AH} & 47.8 & 153.9 & 138.5 & 152.3 & 154.2 & 133.6 & 149.2 & 150.3 \\
\textit{6L\_384D} & 59   & 408.4 & 387.1 & 403.7 & 408.7 & 379.2 & 389   & 397.7 \\ \bottomrule
\end{tabular}
\caption{\label{tab:speed} The number of samples processed by each model per second.}
\end{center}
\end{table*}
\subsection{Aggregating Quantitative Results}
We aggregated results from Tables \ref{tab:scores} and \ref{tab:speed} to be able to design a distillation recipe based on quality and speed gains of student models, e.g the loss for both 3L and 6L\_384D is comparable but considering the faster response time, 6L\_384D seems to be a better alternative. We incorporated our findings and summarized them in Table \ref{tab:min}, which aims to facilitate selecting the right KD configuration in different applications. The \textcolor{green}{\cmark}, \textcolor{red}{\xmark}, and \textcolor{gray}{\textbf{?}} signs in the table indicate whether KD is worthwhile or not. For example, [Layer][CoLA] = \textcolor{green}{\cmark} means no matter which layer-reduction strategy we use, the loss we get from KD is almost consistent, thus we can lean towards a higher compression rate by eliminating more layers and distillation is worthwhile. On the contrary, \textcolor{red}{\xmark} \hspace{0.5mm}shows that the KD approach is harmful. Unlike the two aforementioned signs, \textcolor{gray}{\textbf{?}} is a sign of uncertainly, meaning KD may or may not succeed and it depends on the setup. 

Comparing these factors side-by-side could justify whether KD is a reasonable option or not, i.e  from Table \ref{tab:min}, STSB is immune to all configuration modifications but the attention heads. That means, we can safely reduce the number of layers or shrink the hidden layers yet expect similar high-quality results. From a response-time perspective (shown with \textit{Speed} in the table), it is also worth considering KD for STSB. However, modifying the number of attention heads can yield some inconsistencies, so we would consider KD for STSB only when it is possible to keep the number of attend heads untouched. The table only shows results for CoLA, RTE, and STSB. For the other five tasks, the signs are \textcolor{green}{\cmark} across all factors.
\begin{table}[th]
\begin{center}
\begin{tabular}{c c c c}
\toprule
& CoLA & RTE & STSB \\\midrule
 \textit{Layer} & \textcolor{green}{\cmark} & \textcolor{gray}{\textbf{?}}  & \textcolor{green}{\cmark} \\ 
 \textit{Att. Head} & \textcolor{gray}{\textbf{?}}  & \textcolor{red}{\xmark} & \textcolor{gray}{\textbf{?}}  \\
 \textit{Hidden Dim} & \textcolor{red}{\xmark} & \textcolor{red}{\xmark} & \textcolor{green}{\cmark}\\
\textit{Speed}&\textcolor{gray}{\textbf{?}}  & \textcolor{green}{\cmark} & \textcolor{green}{\cmark}\\\bottomrule
\end{tabular}
\caption{\label{tab:min} Scenarios in which KD could be lossy.}
\end{center}
\end{table}
\section{Conclusion and Future Work}\label{conclusion}
In this paper, we designed a set of experiments to better understand what sort of information is likely to be lost during the distillation process. We focused on different factors, such as, the number of layers or the width of each layer, and showed how they affect the final performance. Our observations showed that in general, the number of layers is not that sensitive to KD, as long as there is no drastic reduction. The width of the hidden layers, on the other hand, showed the highest sensitivity. This feature seems to be quite fragile, and once changed, the final performance changes proportionally. The number of attention heads also shows high sensitivity. We could gain some insight about the behaviour of distilled models via our experiments, but we (as the community) are still far from decoding the problem and deriving conclusions. In our future work, we plan to run more experiments with different teachers, and expand our experiments' domain by testing more datasets and tasks. We are in the era of large language models, and smaller models do not receive enough attention. However, they still have their applications, and  most of real-world problems can be solved effectively and efficiently by the right choice of such compact models. 

%%%%%%%%%%%%%%%%%%%%%%%%%%%%%%%%%%%%%%%%%%%%%%%%%%%%%%%%%%%%
\bibliography{neurips_2021}
\bibliographystyle{neurips_2021}
%%%%%%%%%%%%%%%%%%%%%%%%%%%%%%%%%%%%%%%%%%%%%%%%%%%%%%%%%%%%
%%%%%%%%%%%%%%%%%%%%%%%%%%%%%%%%%%%%%%%%%%%%%%%%%%%%%%%%%%%%
%%%%%%%%%%%%%%%%%%%%%%%%%%%%%%%%%%%%%%%%%%%%%%%%%%%%%%%%%%%%
\newpage
\section{Appendix}
\textbf{Student Models:} \hspace{3mm} Table \ref{KDconfig} provides detailed information about the student models and their configuration.
\begin{table}[th]
\begin{center}
\begin{tabular}{m{12mm} m{11mm} m{11mm} m{13mm} m{1.2cm} }
\toprule
Model & Layer & Attention Heads & Layer Dimension. & Parameter\\ \midrule
Teacher & 12 & 12  & 768 & 124M \\ 
9L & 9  & 12  & 768  & 103M \\ 
6L    & 6  & 12 & 768   & 82M  \\ 
3L    & 3 & 12   & 768 & 61M   \\ 
8AH    & 12   & 8  & 768 & 124M  \\ 
4AH & 12   & 4  & 768  & 124M  \\ 
516D & 12   & 12  & 516   & 65M \\ 
384D & 12   & 12  & 384   & 41M \\
6L\_384D & 6   & 12  & 384   & 30M \\
\bottomrule
\end{tabular}
\caption{\label{KDconfig} Distillation configurations used in our experiments. RoBERTa base is the teacher model. The last columns shows the number of total parameters of each model.}
\end{center}
\end{table}

\textbf{Datasets:} \hspace{3mm} 
Table \ref{tab:glue} provides detail information about  experimental datasets. 
\begin{table}[th]
\begin{center}
\begin{tabular}{ l  l l  l l  l l }
\toprule
Corpus & Task & Train & Dev\\\midrule
 MRPC &  Paraphrase & 3.7k & 408\\
 RTE & NLI & 2.5k &276 \\
 QNLI & QA/NLI &108k & 5.7k \\ 
 CoLA & Acceptability & 8.5K & 1K \\
 MNLI & NLI & 393k &20k \\
 SST2 & Sentiment & 67k & 872 \\
 QQP & Paraphrase & 364k &40k \\
 STSB & Textual Similarity & 5.7K &1.5K \\
 \bottomrule
\end{tabular}
\caption{\label{tab:glue} The statistics of the tasks/datasets from the GLUE collection \citep{DBLP:journals/corr/abs-1804-07461}. Apart from STSB which is a regression task and MNLI which has 3 classes, all the other tasks fall under binary classification.}
\end{center}
\end{table}

\textbf{Qualitative Analysis:} \hspace{3mm} In addition to the quantitative experiments reported in the paper, we ran a set of qualitative analyses. We mainly extracted four types of instances from our datasets where \textit{i}) the teacher and all student models match the true label (TL); \textit{ii}) the teacher and only a subset of performant students agree with TL; \textit{iii}) only the teacher is able to produce TL, and finally \textit{iv}) both the teacher and students fail to match TL. In total, we picked 140 instances and manually analyzed them, in the hope of finding common patterns in these instances. Table \ref{tab:sample} shows a subset of our samples. 
\begin{table}[th]
\begin{center}
\begin{tabular}{m{8mm} m{60mm} m{5mm} m{5mm} m{5mm} m{5mm} m{6mm} m{6mm} m{5mm}}
\toprule
Dataset & Example & TL & $\mathcal{T}$ & 9L & 6L & 516D & 384D & 3L \\\midrule
CoLA & Dana walked and Leslie ran & 1 & 1 & 1 & 1 & 1 & 1 & 1 \\\hline
\multirow{2}{*}{QQP} & \textbf{Q1}: How many zeroes are there in one billion? &\multirow{2}{*}{0} &\multirow{2}{*}{0} &\multirow{2}{*}{0} &\multirow{2}{*}{0} &\multirow{2}{*}{0} &\multirow{2}{*}{0} & \multirow{2}{*}{0}\\ 
 & \textbf{Q2}: How many zeroes does a Googol have? &  &  &  &  &  &  & \\\hline
\multirow{2}{*}{RTE} & \textbf{S1}: Two British soldiers have been arrested in the southern Iraq city of Basra, sparking clashes outside a police station where they are being held & \multirow{2}{*}{1} & \multirow{2}{*}{1} & \multirow{2}{*}{1} & \multirow{2}{*}{1} & \multirow{2}{*}{1} & \multirow{2}{*}{1} & \multirow{2}{*}{\textcolor{red}{0}}  \\
& \textbf{S2}: Two British tanks, sent to the police station where the soldiers are being held, were set alight in clashes & &  &  &  &  &  &\\\hline
\multirow{2}{*}{QNLI} & \textbf{Q}: What religion is the western region mostly? & \multirow{2}{*}{1}& \multirow{2}{*}{1}& \multirow{2}{*}{1} & \multirow{2}{*}{1} & \multirow{2}{*}{\textcolor{red}{0}}  & \multirow{2}{*}{\textcolor{red}{0}} & \multirow{2}{*}{\textcolor{red}{0}}\\
& \textbf{A}: The upper part of Kenya's Eastern Region is home to 10$\%$ of the country's Muslims, where they constitute the majority religious group. &  &  & & & & &\\\hline
\multirow{2}{*}{QQP} & \textbf{Q1}: Which is the best laptop below rs60000? & \multirow{2}{*}{0} & \multirow{2}{*}{0} & \multirow{2}{*}{\textcolor{red}{1}} & \multirow{2}{*}{\textcolor{red}{1}} & \multirow{2}{*}{\textcolor{red}{1}} & \multirow{2}{*}{\textcolor{red}{1}} & \multirow{2}{*}{\textcolor{red}{1}}\\
& \textbf{Q2}: Which is the best laptop to buy under 50k? &  &  & & &  & &\\\hline
  
\multirow{2}{*}{MRPC} & \textbf{S1}: While dioxin levels in the environment were up last year , they have dropped by 75 percent since the 1970s, said Caswell & \multirow{2}{*}{0} & \multirow{2}{*}{0} & \multirow{2}{*}{\textcolor{red}{1}} & \multirow{2}{*}{\textcolor{red}{1}} & \multirow{2}{*}{\textcolor{red}{1}} & \multirow{2}{*}{\textcolor{red}{1}} & \multirow{2}{*}{\textcolor{red}{1}}\\
& \textbf{S2}: The Institute said dioxin levels in the environment have fallen by as much as 76 percent since the 1970s &  &  & &  & &  & \\\hline
\multirow{2}{*}{RTE} & \textbf{S1}: Today's  best estimate of giant panda numbers in the wild is about 1,100 individuals living in up to 32 separate populations mostly in China's Sichuan Province, but also in Shaanxi and Gansu provinces & \multirow{2}{*}{1} & \multirow{2}{*}{\textcolor{red}{0}} & \multirow{2}{*}{\textcolor{red}{0}} & \multirow{2}{*}{\textcolor{red}{0}} & \multirow{2}{*}{\textcolor{red}{0}} &\multirow{2}{*}{\textcolor{red}{0}} & \multirow{2}{*}{\textcolor{red}{0}} \\
&\textbf{S2}: There are 32 pandas in the wild in China & &  & & & &  &  \\\bottomrule
\end{tabular}
\caption{\label{tab:sample} A subset of samples studied in our qualitative analysis. TL stands for True Label and $\mathcal{T}$ is the teacher prediction.}
\end{center}
\end{table}

For the last group where both the teacher and all students fail to detect the correct class we could not find any explainable logic and it is not quite clear why even a strong teacher model cannot learn the task. However, for other categories we noticed some commonalities across different datasets. For trivial examples, all the teacher and student models easily solve the task because we believe \textit{the input is compatible with the nature of the task}, i.e. in the CoLA example provided in the table, the task is to check the grammatical correctness of the input and the short length of it makes the job relatively straightforward. Similarly, in the second sample from QQP, the words ``billion'' and ``Googol'' carry strong signals which makes comparison of Q1 and Q2 easy.

For slightly complex examples where sentences are \textit{long} or the wording varies in the input tuples,  only student models with a higher learning capacity can succeed. The higher the capacity of the student, the higher its chance to mimic its teacher, e.g in the RTE example only the weakest model (3L) fails to predict the right class. Occurrence of common phrases such as ``Two British'', ``clashes'', or ``police station'' also makes compering S1 and S2 less complicated and students equipped with better memory/learning modules can benefit from it. In the case of QNLI the situation is slightly more challenging. The way the same concept is phrased is different from Q1 to Q2 thus only (really) high-capacity models are able to judge correctly. Also, in this scenario, the very same short-length feature that was useful in CoLA hurts the QNLI model as a longer sequence (in Q) could provide the model with more context.    

What we summarized here, clearly, does not universally apply to all KD models and datasets, and it is just our observation but it was interesting to see how there may exit a \textit{common pattern} in the failed/successful cases. Moreover, we found out that regardless of the dataset, model architecture, and task, existing NLP models as well as KD techniques need serious (if not revolutionary) modifications. We encountered examples which are trivial to comprehend for us (as humans) but our best models failed to tackle. The last RTE example in the table could be one of them. Finally, we witnessed multiple cases of \textit{memorization} rather than  \textit{learning}, the well-known shortcoming of neural models, where by changing a single word in a long sentence the behaviour of the model changes completely (as soon as the model is exposed to an unfamiliar input it fails to respond).   
%%%%%%%%%%%%%%%%%%%%%%%%%%%%%%%%%%%%%%%%%%%%%%%%%%%%%%%%%%%%

\end{document}